\documentclass[submission,copyright,creativecommons]{eptcs}
\providecommand{\event}{ICLP/LPNMR-DC 2024}

\usepackage{underscore}
\usepackage[utf8]{inputenc}
\usepackage{amsmath}
\usepackage{amssymb}
\usepackage{microtype}
\usepackage{url}

\providecommand{\sysfont}{\textit}

\newcommand{\clingcon}{\sysfont{clingcon}}
\newcommand{\clingo}{\sysfont{clingo}}
\newcommand{\clingodl}{\clingoM{dl}}
\newcommand{\clingolp}{\clingoM{lp}}

\newcommand{\dingo}{\sysfont{dingo}}
\newcommand{\dlvhex}{\sysfont{dlvhex}}

\newcommand{\ezsmt}{\sysfont{ezsmt}}

\newcommand{\fclingo}{\sysfont{fclingo}}

\newcommand{\mingo}{\sysfont{mingo}}

\newcommand{\clingoM}[1]{\clingo{\small\textnormal{[}\textsc{#1}\textnormal{]}}}

\providecommand{\logfont}{\textrm}

\newcommand{\HT}{\ensuremath{\logfont{HT}}}

\newcommand{\HTC}{\ensuremath{\logfont{HT}_{\!c}}}

\newcommand{\tuple}[1]{\ensuremath{\langle #1 \rangle}}

\newcommand{\handt}{\tuple{H,T}}

\newcommand{\HTLB}{\ensuremath{\logfont{HT}_{\logfont{LB}}}}
\newcommand{\ASPAC}{\ensuremath{\mathit{ASP}({\mathcal{AC}})}}

\newcommand{\CoomFont}[1]{\textsc{#1}}
\newcommand{\coom}{\CoomFont{Coom}}
\newcommand{\Coom}{\coom}

\newcommand{\coomsuite}{\CoomFont{CoomSuite}}

\newcommand{\potassco}{\textsc{Potassco}}

\newcommand{\code}[1]{\ensuremath{\mathtt{#1}}} 

\newcommand\diffc[3]{\code{\&diff\{\mathit{#1}-\mathit{#2}\} \ \mbox{\tt <=} \ \mathit{#3}}}

\title{Hybrid Answer Set Programming: Foundations and Applications}

\author{%
  Nicolas R\"uhling \\
  \institute{University of Potsdam, Germany \\ \textit{Institute of Computer Science} \\ \textit{An der Bahn 2, 14476 Potsdam}}
  \email{nruehling@uni-potsdam.de}
}

\def\titlerunning{Hybrid ASP: Foundations and Applications}
\def\authorrunning{N. R\"uhling}

\hypersetup{
  bookmarksnumbered,
  pdftitle    = {\titlerunning},
  pdfauthor   = {\authorrunning},
  pdfsubject  = {ICLP/LPNMR-DC 2024},               
  pdfkeywords = {Answer Set Programming, Logic of Here-and-there, Configuration} 
}

\begin{document}
\maketitle

\section{Introduction}\label{sec:introduction}
Answer Set Programming (ASP; \cite{lifschitz19a}) is being increasingly applied to solve problems from the real-world.
However, in many cases these problems have a heterogenous nature which requires features beyond the current capacities of solvers like \clingo\ \cite{gekakasc17a}.
Consider for example the field of configuration \cite{kbc14}; one of the early successful applications of ASP \cite{gekasc11c,gescer19a}.
A configuration problem usually consists of (at least) a partonomy where parts are parameterized by attributes whose values in turn are restricted by constraints.
%
While in simple cases these attributes are discrete, many industrial applications require attributes
that range over large numeric domains (eg.\ precisions in the milimeter range might be needed).
Further, calculations over these attributes can be of linear nature (eg.\ calculating the total weight by summing up the weight of all parts) as well as
non-linear (eg.\ area or volume of an object, inclination of a conveyor belt, etc).
Standards ASP solvers like \clingo\ quickly reach their limits when dealing with numeric ranges and calculations
as they need to explicitly ground all possible values.
Apart from this, representing constraints in ASP that go beyond simple arithmetic expressions or aggregations generally requires considerable effort.

Over the last years, hybrid solvers such as
\clingcon\ \cite{bakaossc16a}\footnote{\url{https://potassco.org/clingcon}} and
\clingodl\ \cite{jakaosscscwa17a}\footnote{\url{https://github.com/potassco/clingo-dl}}
which make use of dedicated inference methods for certain kinds of constraints over finite integer domains
have already been successfully applied to many problems such as train scheduling \cite{abjoossctowa21a} and warehouse delivery \cite{rascwachliso23a}.
However, what is still missing, is a solid, semantic underpinning of these systems. 

This issue has first been addressed by introducing the Logic of \emph{Here-and-There with constraints} (\HTC; \cite{cakaossc16a}) as
an extension of the Logic of \emph{Here-and-There} (\HT; \cite{heyting30a}) and its non-monotone extension \emph{Equilibrium Logic} \cite{pearce06a}.
Nowadays, \HT\ serves as a logical foundation for ASP and has facilitated a broader understanding of this paradigm.
The idea is that \HTC\ (and other extensions; see Section~\ref{sec:background}) play an analogous role for hybrid ASP.

There remain many open questions about these logics regarding their fundamental characteristics
as well as their practical use in solvers, ie.\ how they can guide the implementation.
Having a formal understanding of these hybrid logics is also needed
to better understand the inherent structure of the (real-world) problems they are applied to, eg.\ configuration, and to improve their representations in ASP.

\section{Background}\label{sec:background}

\subsection{Hybrid Solvers}\label{sec:systems}
Nowadays, ASP solver \clingo\ supports so-called \emph{theory atoms} which allow for foreign inference methods \cite{jakaosscscwa17a,karoscwa21a}
following the approach of SAT \emph{modulo theories} (SMT; \cite{SATHandbook}).
This has greatly facilitated the development of ASP-based special-purpose systems which make use of dedicated inference methods for
certain subclasses of constraints such as difference logic and linear programming.
The general idea is that some external theory serves as an oracle by certifying some of a program's stable models and has been characterized for \clingo\ in \cite{cafascwa23a}.
We proceed by giving a quick introduction of some of the hybrid solvers that are part of the \potassco\ suite \footnote{\url{https://potassco.org/}}.

The system \clingcon\ is a solver for \emph{Constraint Answer Set Programming} (CASP) and extends the input language of \clingo\ with linear equations,
represented as theory atoms of the form
\begin{align}\label{clingcon:linear:constraint}
    \code{\&sum\{\mathit{k_1*x_1};\dots;\mathit{k_n*x_n}\}} \prec k_0
\end{align}
where $x_i$ is an integer variable and $k_i\in\mathbb{Z}$ an integer constant for $0\leq i\leq n$;
and $\prec$ is a comparison symbol such as \code{<=}, \code{=}, \code{!=}, \code{<}, \code{>}, \code{>=}.
In \clingo, theory predicates are preceded by `\code{\&}'.

System \clingodl\ has a more restricted syntax which allows for \emph{difference constraints} over integers.
This is a subset of the syntax in (\ref{clingcon:linear:constraint}) where theory atoms have the fixed form
$\code{\&sum\{1 * \mathit{x}; (-1)*\mathit{y}\} <= \mathit{k}}$
but are rewritten instead as:
\begin{align}\label{clingodl:difference:constraint}
    \diffc{x}{y}{k}
\end{align}
with $x$ and $y$ integer variables and $k \in\mathbb{Z}$.

A third system is \clingolp\ \footnote{\url{https://github.com/potassco/clingoLP}} which extends \clingo\ to solve linear constraints
as dealt with in Linear Programming (LP).
The syntax is identical to (\ref{clingcon:linear:constraint}) but the domain now ranges over the real numbers.
Notably, \cite{cafascwa23a} contains a formal characterization of all three just mentioned systems.

Lastly, a recent addition is system \fclingo\ \footnote{\url{https://github.com/potassco/fclingo}} which makes use of \clingcon\
to solve ASP modulo conditional linear constraints with founded variables.
While in \clingcon\ all integer variables need to have a value assigned,
\fclingo\ adds a notion of undefinedness and foundedness as known from ASP,
ie.\ there needs to be a justification in the logic program if a variable receives a value in an answer set.
Further, the conditional aspect of the linear constraints can be seen as a generalization of the concept of aggregates commonly used in ASP.
The syntax of \fclingo\ accomodates so-called \emph{assignments} which guarantee that a variable only gets assigned a value if
all other variables in its definition are itself defined, ie.\ justified at some other part in the logic program.
For instance, the expression
\begin{align*}\label{fclingo:assignment}
    \code{\&in\{\mathit{y}..\mathit{y}\} =: \textit{x}}
\end{align*}
only assigns the value of $y$ to $x$ if $y$ has been defined by some other rule.
Omitting the assignment would permit $y$ and $x$ to take arbitrary values if not defined elsewhere, thereby circumventing the principle of foundedness.

Further systems not developed by \potassco\ include ASP solver \dlvhex\ \cite{redl16a} which supports a similar concept of theory atoms as \clingo.
Other CASP systems include \dingo\ \cite{janise09a}, \mingo\ \cite{lijani12a} and \ezsmt\ \cite{liesus16a}.
Different from the aforementioned systems, all three rely on translations to non-ASP solvers.

\subsection{The Logic of Here-and-There and Hybrid Extensions}\label{sec:semantics}
The logics \HT\ and \emph{Equilibrium Logic} nowawadays serve as a logical foundation for (plain) ASP,
having brought upon fundamental results such as the notion of \emph{strong equivalence} \cite{lipeva01a}.
The idea of \HT\ is that of two worlds $h$ and $t$, generally called \emph{here} and \emph{there}.
\footnote{This is based on Kripke semantics for intuitionistic logic, see \cite{dalen86a}}
More precisely, an \HT-interpretation is a pair \handt\ of sets of atoms such that $H \subseteq T$.
This gives rise to a three-valued logic where atoms can either be \textit{true}, \textit{false} or \textit{undefined}.
A formula $\varphi$ is \emph{satisfied} or \emph{holds} in a model \handt\, in symbols $\handt \models \varphi$, if it is true in the model, ie.\ satisfied at the $h$\nobreakdash-world.
A model \handt\ of a theory $\Gamma$ is called an \emph{equilibrium model} if
(i) it is total, ie.\ $H = T$, and
(ii) for any $H'$ such that $H' \subset T$, $\handt \not\models \Gamma$.
The term equilibrium model was coined in \cite{pearce96a} and there is complete agreement between equilibrium models and the stable models of logic programs as defined in \cite{gellif88b}.

In an attempt to provide a solid, logical foundation for hybrid systems such as the ones introduced in Section~\ref{sec:systems},
a number of extensions of \HT\ for incorporating constraints have been introduced.

The Logic of \emph{Here-and-There with constraints} (\HTC; \cite{cakaossc16a}) allows for capturing constraint theories in the non-monotonic setting
and has subsequently been extended with aggregate functions over constraint values and variables \cite{cafascwa20a,cafascwa20b}.
In \cite{cafascwa20b} specifications for aggregate functions in terms of \HTC\ are given based on two different semantic principles.
While the semantics given in \cite{ferraris11a} ensures that aggregate terms are always defined,
\cite{gelzha14a} prohibits so-called \emph{vicious cycles}. 
%
We also refer to the former as \emph{Ferraris} and to the latter as \emph{Gelfond-Zhang} (GZ) aggregate semantics.

The Logic of \emph{Here-and-there with lower bound founded variables} (\HTLB; \cite{cafascsc19a}) generalizes the concept of \emph{foundedness} to integer variables.
The idea is that variables get assigned the smallest integer value that can be justified.
This can be seen as a generalization of plain \HT\ if one regards Boolean truth values as ordered by letting \textit{true} be greater than \textit{false}.

Both of these extensions can be seen as \emph{black-box} approaches in the sense that the constraints are incorporated
as special entities whose syntax and satisfaction relations are generally left open.
Thus, the intricacies of the hybrid part are mostly unknown from the logic program perspective.
Another \HT\ extension with a \emph{white-box} approach of constraints is \ASPAC\ \cite{eitkie20a}
which generalizes logical connectives as a particular case of more general operations on weighted formulas over semirings.
In this setting, operators like logical conjunction $\wedge$ become just one more possible operation that can be combined with others,
such as addition or multiplication (depending on the underlying semiring).
This results in a very expressive and powerful formalism but at the price of a more complex semantics and the requirement of a semiring structure.

Further white-box approaches are based on the incorporation of intensional or non-Herbrand functions in ASP.
For instance, \cite{cabalar11a} added partial intensional functions to a quantified First-Order version of \HT\ \cite{peaval04b} and
later extended this to sets and aggregates \cite{cafafape18a}.

\subsection{Configuration}
A wide range of approaches exist for representing and solving configuration problems
across various paradigms~\cite{junker06a,hofestrybawo14a}.
In recent years, ASP has emerged as a promising alternative,
as evidenced by several applications~%
\cite{%
    gekasc11c,%
    fefaateruraz17a,%
    gescer19a,%
    hebasasc22a%
}.
Moreover, \cite{faryscsh15a} developed an object-oriented approach to configuration by directly defining concepts in ASP.
In the context of interactive configuration,
\cite{fahakrscscta20a}~conducted a comparative evaluation of various systems,
including the ASP solver \clingo\ as well as SAT and CP systems,
for their suitability in this context, finding \clingo\ to be as capable as any other system.

\section{Research}\label{sec:research}

My research focuses on the foundations of hybrid ASP with the goal of both
understanding better its fundamental properties and
exploiting this knowledge to guide and improve solver implementations.
Further, as this research is motivated by problems in real-world applications,
another goal is to better understand the essence of these problems and how they can be solved using hybrid ASP.
More precisely, the objective is to find mathematical or logical formalizations of these problems which subsequently serve as basis for succint but general ASP representations.
These two goals are reciprocally beneficial as a deeper insight into real-world problems will make clearer the necessary research directions on the foundational level.
In the context of applications, my current focus lies on problems in the realm of (industrial product) configuration.

\subsection{Contributions and Future Work}
Regarding the theoretical aspects of my research I am currently working on the theoretical foundations of solver \fclingo\
with the goal of improving the current implementation.
Here, one of the open issues is that current results in \HTC\ only allow for the use of GZ aggregate semantics (see Sec.~\ref{sec:semantics}) in \fclingo.
However, we would like to be able to use Ferraris aggregate semantics which guarantee definedness as known from \clingo.
Our current approach here consists of finding a suitable translation between the two semantics.

Another open issue is the formalization of solver \clingodl\ by means of logic \HTLB.
The concept of assigning a minimal, founded value to integer variables of \HTLB\ seems like a natural match with
the difference constraints in \clingodl\ which are defined as inequalities, thus, generally have multiple valid solutions
but only one or a few minimal ones.
%
%

On the practical side of my research, preliminary results have been found in application of (plain and hybrid) ASP to configuration problems.
In \cite{ruscst23a} we developed a principled approach to configuration that included a mathematical formalization
of configuration problems with an ASP-based solution.
We defined a configuration problem in terms of an abstract model and a concrete instantiation.
While the model serves as a blueprint for all possible configurations, the instantation represents a solution.
This work was accompanied by a corresponding fact format and two ASP encoding, one for \clingo\ and one for \fclingo,
which were subsequently made public \footnote{\url{https://github.com/potassco/configuration-encoding}}.

A similar but slightly different work has been done in \cite{baheosreruscwa24a} where we developed the \coomsuite\ \footnote{\url{https://github.com/potassco/coom-suite}},
a workbench for experimentation with industrial-scale product configuration problems.
The \coomsuite\ is built around product configuration language \coom \cite{coomlang}
\footnote{\Coom\ is a domain-specific language developed by denkbares GmbH and used in numerous industrial applications}
and provides a \coom\ grammar for parsing, a specialized ASP translator for conversion into facts, two encodings (one for \clingo\ and one for \fclingo)
as well as various benchmark sets.
The intention is to ease the development of powerful methods able to perform in industrial settings.

Future work here includes the further study of suitable representations for hybrid solver \fclingo.
The current \fclingo\ encodings do not necessarily use all features the solver has to offer, eg.\ undefinedness of numeric variables,
but rather leaves this to non-hybrid ASP.
The reason for this is that these encodings have been constructed with a plain ASP encoding as base,
only modifying the necessary parts.
An approach we want to pursue here is to find a logical formalization of configuration problems in terms of \HTC\
and use this as basic for new encodings which make more natural use of \fclingo's features.
We expect that this will not only improve the knowledge representation but also the performance of the solver.

\bibliographystyle{eptcs}

\end{document}